\documentclass{article}

 \usepackage[dblblindworkshop, preprint]{neurips_2025}
\workshoptitle{Imageomics}
\usepackage[utf8]{inputenc} % allow utf-8 input
\usepackage[T1]{fontenc}    % use 8-bit T1 fonts
\usepackage{hyperref}       % hyperlinks
\usepackage{url}            % simple URL typesetting
\usepackage{booktabs}       % professional-quality tables
\usepackage{amsfonts}       % blackboard math symbols
\usepackage{nicefrac}       % compact symbols for 1/2, etc.
\usepackage{microtype}      % microtypography
\usepackage{xcolor}         % colors
\usepackage{subcaption} 
\usepackage{graphicx}
\usepackage{float}
\usepackage{amsmath}

\title{Bi-Encoder Contrastive Learning for Fingerprint and Iris Biometrics}

\author{%
  Matthew So \\
  Department of Computer Science \\
  Columbia University \\
  \texttt{ms5513@columbia.edu}
  \And
  Judah Goldfeder \\
  Department of Computer Science \\
  Columbia University \\
  \texttt{jag2396@columbia.edu}
  \And
  Mark Lis \\
College of Medicine \\
SUNY Downstate Health Sciences University \\
Mark.Lis@downstate.edu \\
  \And
  Hod Lipson \\
  Department of Mechanical Engineering \\
  Columbia University \\
  \texttt{hod.lipson@columbia.edu}
}

\begin{document}
\maketitle
\begin{abstract}
There has been a historic assumption that the biometrics of an individual are statistically uncorrelated.  We test this assumption by training Bi‑Encoder networks on three verification tasks, including fingerprint-to-fingerprint matching, iris-to-iris matching, and cross-modal fingerprint-to-iris matching using 274 subjects with $\sim$100k fingerprints and 7k iris images.  We trained ResNet‑50 and Vision Transformer backbones in Bi-Encoder architectures such that the contrastive loss between images sampled from the same individual is minimized.  The iris ResNet architecture reaches 91 ROC AUC score for iris-to-iris matching, providing clear evidence that the left and right irises of an individual are correlated.  Fingerprint models reproduce the positive intra‑subject suggested by prior work in this space. This is the first work attempting to use Vision Transformers for this matching.  Cross‑modal matching rises only slightly above chance, which suggests that more data and a more sophisticated pipeline is needed to obtain compelling results.  These findings continue challenge independence assumptions of biometrics and we plan to extend this work to other biometrics in the future. Code available: \url{https://github.com/MatthewSo/bio_fingerprints_iris}
\end{abstract}

\section{Introduction}
\label{sec:intro}

Biometric traits such as fingerprints, voice, iris texture, hand geometry, and face have a history of being assumed to be statistically independent when sampled from a single individual. Even modern security systems and forensic workflows explicitly encode this assumption in their decision rules and probabilistic models \citep{1262028,HOLLINGSWORTH20111493,doi:10.1126/science.1738844}. Recent evidence, however, questions this assumption. Guo \emph{et~al.} reported statistically significant intra-subject patterns in fingerprints and proposed a ResNet-style architecture that predicts same subject fingerprint pairs with better than random chance \citep{doi:10.1126/sciadv.adi0329}. This result has generated substantial interest across the biometrics community \citep{Starr2024Fingerprints, Baily2024FingerprintSimilarity}.

A similiar independence assumption has persisted in the iris recognition space. Daugman’s foundational work framed the left and right irises of an individual as fundamentally random and uncorrelated \citep{doi:10.1098/rspb.2001.1696}, and Hollingsworth \emph{et~al.} reaffirmed that the left and right irises of the same person have uncorrelated irises when they assert, ``it is generally accepted that the left and right irises of the same person have uncorrelated iris codes''
 \citep{HOLLINGSWORTH20111493}.  In \citep{fangiris}, researchers experiment with matching genetically identical irises using grayscaled versions of the images on different channels of a CNN. They were able to demonstrate the predictive power of CNNs to match the irises of matching vs non-matching individuals.

In this paper we extend the empirical investigation of intra subject biometric dependence in three key ways:
\begin{enumerate}
\item We attempt to replicate the fingerprint study of Guo \emph{et~al.} under a low-data regime, while also evaluating additional deep learning architectures.
\item We conduct an analysis of potential correlations between an individual’s two irises using contrastive loss with bi-encoder networks.
\item We examine cross-biometric correlation, and propse a framework to learn the extent to which patterns in fingerprints and irises can indicate a shared origin.
\end{enumerate}

Together, these efforts are aimed toward the development of a more nuanced understanding of biometric uniqueness. The work here is a jumping-off point for further research that will have direct implications for the design and security analysis of next generation biometric systems.

%-------------------------------------------------------------------------

\section{Problem Formulation and Approach}
\label{sec:methodology}

\label{sec:problem}

Fundamentally, the problem we are trying to solve is determining if two images (two fingerprints, two irises, one iris/one fingerprint) come from the same individual or come from two different individuals. 
Formally, we treat biometric verification as a binary decision task.  
Given two images \(x_1 \in \mathcal{X}_1\), \(x_2 \in \mathcal{X}_2\) ($\mathcal{X}_1$ may be the same, in the case of iris-to-iris matching, or different, in the case of fingerprint-to-iris matching, from $\mathcal{X}_2$),  
the goal is to estimate  

\[
f : \mathcal{X}_{1}\times\mathcal{X}_{2} \;\longrightarrow\; \{0,1\},
\]

where \(f(x_1,x_2) = 1\) denotes a valid pair and \(0\) an invalid pair.  
Bi-Encoder (also known as Siamese networks) networks are excellent model candidates for this type of matching problem on images, and a similar approach was taken in Guo et al \citep{10182466}. 

In classic Bi-Encoder architectures, we train a network that takes in two images (generally in the same representational space) and learns embeddings of the images \citep{Koch2015SiameseNN}. During training, one calculates the loss using a Contrastive Loss function that calculates a distance metric between the embeddings and then penalizes matching pairs with high distances and non-matching pairs with low distances. The contrastive loss in our formulation is calculated as:

\begin{align*}
\mathcal{L}(x_1,x_{2},y) = (1-y) \,\lVert g(z_{1})-g(z_{2}) \rVert_{2}^{2} +  
y\,\bigl[\max(0,\,m-\lVert g(z_{1})-g(z_{2})\rVert_{2})\bigr]^{2}
\end{align*}

where $g$ is the embedding network, $m$ is the margin and $y$ is the true label for $x_1$ and $x_2$

This results in embeddings that should be similar for matching pairs and far apart for non-matching pairs. 
% Fig. \ref{fig:classifc_Bi-Encoder_network} visualizes this architecture.

% \begin{figure}[h]
%     \centering
%     \includegraphics[width=1\linewidth]{siamese.png}
%     \caption{Classic Bi-Encoder Network}
%     \label{fig:classifc_Bi-Encoder_network}
% \end{figure}

In attempting to create networks to predict whether two random irises belong to the same individual (iris-to-iris) or two random fingerprints belong to the same individual (fingerprint-to-fingerprint), we used a traditional Bi-Encoder architecture. In tackling the one iris and one fingerprint problem, our first attempt involved a naive Bi-Encoder/Two Tower hybrid network approach that involved training two networks simultaneously, both of which had not previously seen fingerprint or iris data during pre-training. We also attempted to use the pretrained Bi-Encoder models from the iris-to-iris and fingerprint-to-fingerprint specific task models. 

% This approach can be seen in Fig. \ref{fig:naive_Bi-Encoder_network-label}

% \begin{figure} [h]
%     \centering
%     \includegraphics[width=1\linewidth]{naive model.png}
%     \caption{NFingerprint/Iris Approach Architecture}
%     \label{fig:naive_Bi-Encoder_network-label}
% \end{figure}

% After limited success with this approach, 

% Fig. \ref{fig:advanced_Bi-Encoder_network-label} below shows the attempted network surgery.

% \begin{figure}[h]
%     \centering
%     \includegraphics[width=1\linewidth]{advanced.png}
%     \caption{Advanced Pre-trained Fingerprint Iris Approach}
%     \label{fig:advanced_Bi-Encoder_network-label}
% \end{figure}

\section{Methodology}\label{sec:formatting}

\subsection{Dataset}\label{sec:format}
We use a de-identified multi-biometric dataset collected by West Virginia University and Clarkson University, which links records via subject identifiers and enables cross-trait analysis while preserving privacy \citep{dataset}. For this study we selected subjects with both fingerprint and iris data, yielding 274 individuals with $\sim$100k fingerprint images and 7k iris images.

\subsection{Data Preparation}\label{sec:data_preparation}
All images were grouped by subject and split at the subject level into \textbf{train}/\textbf{val}/\textbf{test} sets (80\%/10\%/10\%), preventing identity leakage across splits. For each task we constructed balanced (1:1) positive/negative pair sets within each split:
\begin{itemize}
\item \textbf{Iris–iris:} all left–right iris combinations from the same subject (label $1$); negatives were left–right pairs from different subjects in the same split (label $0$).
\item \textbf{Fingerprint–fingerprint:} pairs of different fingers from the same subject (label $1$); negatives drawn from different subjects (label $0$).
\item \textbf{Iris–fingerprint:} each iris of a subject paired with each fingerprint of the same subject (label $1$); negatives paired with fingerprints from randomly sampled other subjects (label $0$).
\end{itemize}
Because image counts per subject vary, we normalized subject frequency by additional sampling of under-represented identities so that each individual contributes equally to training. When comparing multiple backbones, the exact same subject lists for train/val/test were reused across models to avoid cross-model leakage.

\subsection{Training}\label{sec:training}
We train Bi-Encoder models with ResNet-50 and Vision Transformer backbones \citep{DBLP:RESNET50,DBLP:ViT}. The final classification layer is replaced by a projection head (fully connected + batch normalization) producing the embedding. Models are optimized with Adam (initial LR $3\times10^{-4}$) and a step-decay scheduler (step size $=5$ epochs, $\gamma=0.1$). We apply light augmentations (color jitter, small rotations, normalization) during training and disable augmentation at test time. Early stopping is based on validation performance, and decision thresholds for binary classification are selected on the validation set.

\begin{table}[t]
  \centering
  \caption{Principal training configs. All models use ImageNet-1K initialization; $^\dagger$ indicates towers pretrained from i1/f1.}
  \label{tab:hyperparams}
  \small
  \begin{tabular}{@{}lllrrr@{}}
    \toprule
    ID & Task & Backbone & Batch & GPUs & LR \\
    \midrule
    i1 & Iris–Iris              & RN50        & 50 & 7 & 3e-4 \\
    i2 & Iris–Iris              & ViT-B/16    & 16 & 3 & 3e-4 \\
    i3 & Iris–Iris              & ViT-L/32    & 16 & 2 & 3e-4 \\
    f1 & Fingerprint–Fingerprint & RN50        & 50 & 7 & 3e-4 \\
    f2 & Fingerprint–Fingerprint & ViT-B/16    & 16 & 3 & 3e-4 \\
    c1 & Iris–Fingerprint       & 2$\times$RN50 & 50 & 7 & 3e-4 \\
    c2 & Iris–Fingerprint       & 2$\times$RN50$^\dagger$ & 50 & 7 & 5e-5 \\
    \bottomrule
  \end{tabular}

  \vspace{2pt}
  \footnotesize $^\dagger$ Towers initialized from i1 (iris) and f1 (fingerprint).
\end{table}

\section{Results and Discussion}\label{sec:results}

We evaluate three verification tasks using ROC AUC as the primary metric and report complementary accuracy, precision, and recall (Fig.\ref{fig:roc_auc},\ref{fig:accuracy},\ref{fig:precision},\ref{fig:recall}).

\subsection{ROC AUC Performance}\label{sec:roc_results}
\begin{figure}[t]
  \centering
  \includegraphics[width=0.95\linewidth]{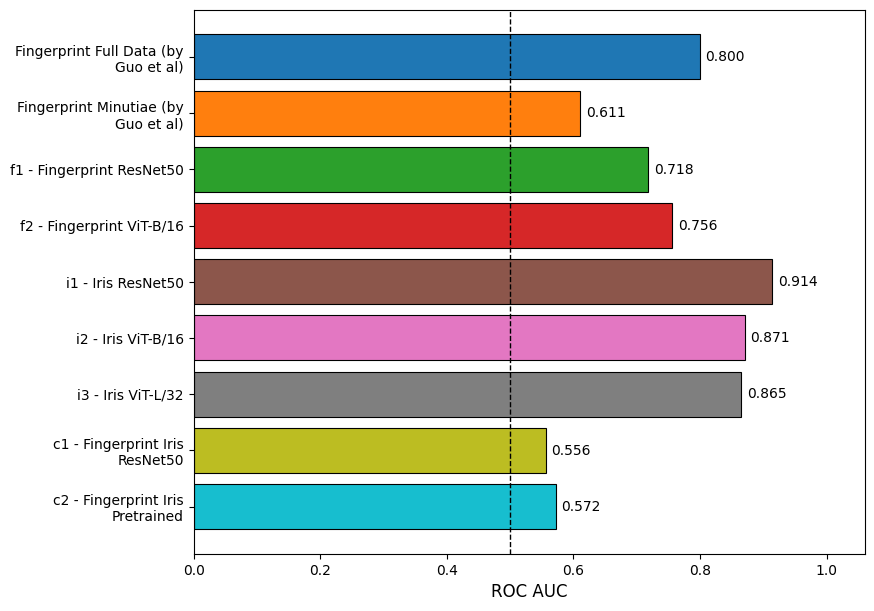}
  \caption{ROC AUC (including Guo \emph{et~al.} results).}
  \label{fig:roc_auc}
\end{figure}

Figure~\ref{fig:roc_auc} aggregates our ROC AUC scores alongside two fingerprint baselines from Guo \emph{et~al.} \citep{doi:10.1126/sciadv.adi0329}. Overall, iris-iris achieves the strongest separability; fingerprint-fingerprint reproduces the positive intra-subject signal from prior work; cross-modal iris-fingerprint remains weak.

% --- Figure 2: Triangle layout for Accuracy / Precision / Recall ---

% \begin{figure}[h]
%   \centering
%   % Top row
%   \begin{subfigure}{0.48\linewidth}
%     \centering
%     \includegraphics[width=\linewidth]{roc_auc_updated2.png}
%     \caption{ROC AUC (including Guo \emph{et~al.} results)}
%     \label{fig:roc_auc}
%   \end{subfigure}\hfill
%   \begin{subfigure}{0.48\linewidth}
%     \centering
%     \includegraphics[width=\linewidth]{accuracy_updated2.png}
%     \caption{Accuracy}
%     \label{fig:accuracy}
%   \end{subfigure}

%   \vspace{0.4em}

%   % Bottom row
%   \begin{subfigure}{0.48\linewidth}
%     \centering
%     \includegraphics[width=\linewidth]{precision_updated2.png}
%     \caption{Precision}
%     \label{fig:precision}
%   \end{subfigure}\hfill
%   \begin{subfigure}{0.48\linewidth}
%     \centering
%     \includegraphics[width=\linewidth]{recall_updated2.png}
%     \caption{Recall}
%     \label{fig:recall}
%   \end{subfigure}

%   \caption{Summary of model performance on held-out test sets.}
%   \label{fig:metrics}
% \end{figure}

\subsection{Iris-to-Iris}\label{sec:iris_results}
The ResNet-50 bi-encoder attains ROC AUC $0.91$, indicating that contralateral irises are not statistically independent. Qualitative inspection of errors suggests many false negatives stem from poor image quality, eyelid/eyelash occlusion, or specular reflections. Vision Transformers (ViT-B/16, ViT-L/32) exceed chance but plateau in the mid-to-high 0.80s; with limited data, CNN inductive biases appear advantageous relative to ViTs trained from scratch.

\subsection{Fingerprint-to-Fingerprint}\label{sec:fp_results}
We reproduce the qualitative conclusion of Guo \emph{et~al.} that fingerprints from the same subject exhibit intra-subject signal, though our ROC AUC does not match their best report. Likely factors include fewer participants ($\sim$3$\times$ fewer), less extensive pretraining (e.g., synthetic data), and architectural/training differences. Notably, ViT-B/16 outperforms ResNet-50 in our setting, suggesting transformer backbones can be competitive when enough fingerprint data are available.

\subsection{Fingerprint-to-Iris (Cross-Modal)}\label{sec:cross_results}
Cross-modal verification yields only slight gains over chance, even with our strongest two-tower configurations. This suggests a weaker measurable correlation between fingerprint and iris traits under our data and training regime; larger datasets, cross-modal pretraining, or stronger alignment objectives may be required to uncover robust signal.

\section{Interpretability}
\label{sec:interpretability}

To explore the internal representations learned by the networks, we adopted a DeepDream‑style  open‑source CNN visualisation code of Ozbulak \citep{uozbulak_pytorch_vis_2022} as was done by Guo \emph{et~al.}. For select CNN layers, we perform gradient ascent on an initial image comprised of noise to see what features in the input image the layer is responding to. This gives intuitive insight into the visual patterns that affect the results.

\begin{figure}[h]
  \centering

  % Top row: Fingerprint
  \begin{subfigure}[t]{0.48\linewidth}
    \centering
    \includegraphics[width=0.47\linewidth]{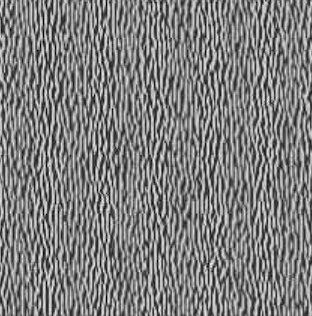}\hfill
    \includegraphics[width=0.48\linewidth]{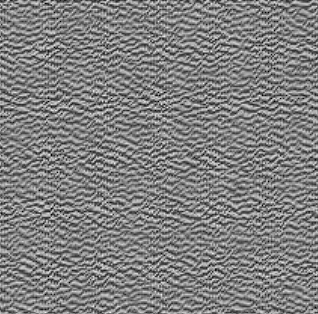}
    \caption{DeepDream visualisations from \textbf{Layer 1} of the fingerprint ResNet-50. The results resemble course fingerprint patterns.}
    \label{fig:fp_layer1}
  \end{subfigure}\hfill
  \begin{subfigure}[t]{0.48\linewidth}
    \centering
    \includegraphics[width=0.48\linewidth]{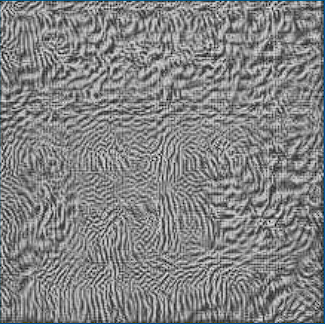}\hfill
    \includegraphics[width=0.485\linewidth]{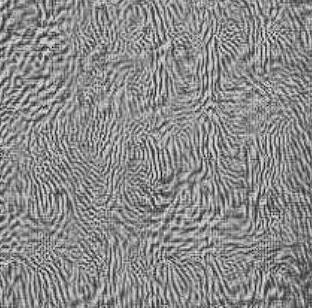}
    \caption{DeepDream visualisations from \textbf{Layer 4} of the fingerprint ResNet-50. Later filters capture complex minutiae-like ridge patterns.}
    \label{fig:fp_layer4}
  \end{subfigure}

  \vspace{0.8em}

  % Bottom row: Iris
  \begin{subfigure}[t]{0.48\linewidth}
    \centering
    \includegraphics[width=0.46\linewidth]{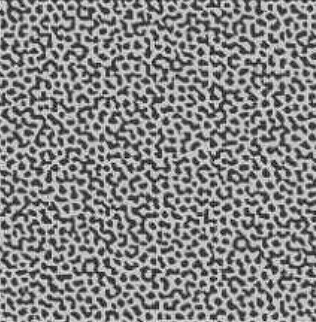}\hfill
    \includegraphics[width=0.47\linewidth]{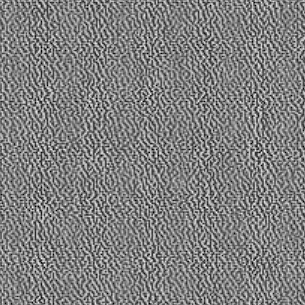}
    \caption{DeepDream visualisations from \textbf{Layer 1} of the iris ResNet-50. Filters the core patterns of irises.}
    \label{fig:iris_layer1}
  \end{subfigure}\hfill
  \begin{subfigure}[t]{0.48\linewidth}
    \centering
    \includegraphics[width=0.49\linewidth]{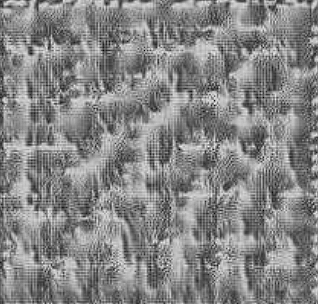}\hfill
    \includegraphics[width=0.49\linewidth]{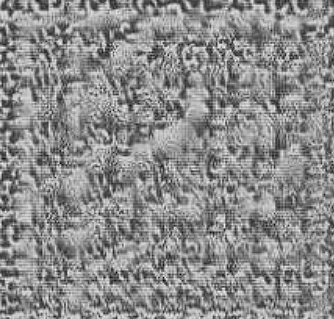}
    \caption{DeepDream visualisations from \textbf{Layer 4} of the iris ResNet-50. Later filters seem to respond to complex patterns found in irises.}
    \label{fig:iris_layer4}
  \end{subfigure}

\end{figure}

\paragraph{Early‑layer features.}
Figures \ref{fig:fp_layer1} and \ref{fig:iris_layer1} present the resultant images for the first convolutional block of the fingerprint and iris ResNet50 architectures, respectively.  The networks for both tasks seem to be learning the core structures of fingerprints and irises.  

\paragraph{Deep‑layer features.}
In deeper layers (Figures \ref{fig:fp_layer4} and \ref{fig:iris_layer4}), the images contain more complex representations of the patterns found on fingerprints and irises.  For fingerprints, the network appears to focus on minutiae‑like patterns, including swirls, ridges and bifurcations, that may be used to identify an individual.  The iris network, on the other hand, seems to demonstrate patterns of the stoma of the iris. The notable differences between early and late layers suggest that the Bi-Encoder architecture learns a sequential hierarchy from coarse structural features to complex patterns.

% \begin{figure}[h]
%   \centering
%   \includegraphics[width=0.45\linewidth]{Fingerprint - Layer 1 Example.png}\hfill
%   \includegraphics[width=0.45\linewidth]{fingerprint-layer1.png}
%   \caption{DeepDream visualisations from \textbf{Layer 1} of the fingerprint ResNet‑50.  The results resemble course fingerprint patterns.}
%   \label{fig:fp_layer1}
% \end{figure}

% \begin{figure}[h]
%   \centering
%   \includegraphics[width=0.45\linewidth]{fingerprint_lay4.png}\hfill
%   \includegraphics[width=0.45\linewidth]{fingerprint_layer42.png}
%   \caption{DeepDream visualisations from \textbf{Layer 4} of the fingerprint ResNet‑50.  Later filters capture complex minutiae‑like ridge patterns.}
%   \label{fig:fp_layer4}
% \end{figure}

% \begin{figure}[h]
%   \centering
%   \includegraphics[width=0.45\linewidth]{iris_lay1_1.png}\hfill
%   \includegraphics[width=0.45\linewidth]{iris_lay_1_2.png}
%   \caption{DeepDream visualisations from \textbf{Layer 1} of the iris ResNet‑50.  Filters the core patterns of irises.}
%   \label{fig:iris_layer1}
% \end{figure}

% \begin{figure}[h]
%   \centering
%   \includegraphics[width=0.45\linewidth]{iris_lay_4.png}\hfill
%   \includegraphics[width=0.45\linewidth]{iris_lay_4_2.png}
%   \caption{DeepDream visualisations from \textbf{Layer 4} of the iris ResNet‑50.  Later filters seem to respond to complex patterns found in irises.}
%   \label{fig:iris_layer4}
% \end{figure}
\section{Future Work}\label{sec:future}
The main areas of future work are: (i) systematic hyperparameter and architecture search with scaling studies; (ii) automated image-quality scoring and filtering of low-quality samples, complemented by targeted augmentations; (iii) domain-specific pretraining (e.g., synthetic biometrics, self-supervised objectives) and exploration of alternative contrastive losses/margins and alignment objectives; and (iv) broader cross-modal studies on larger, more diverse datasets, including voice, vein patterns, and hand geometry.

%------------------------------------------------------------------------
\section{Conclusion}\label{sec:conclusion}
This work revisits the independence assumption for biometric traits using bi-encoder contrastive learning. Under a tighter data budget, we replicate fingerprint intra-subject effects and, in a large left-right iris study, observe substantial signal (ROC AUC 0.91 with ResNet-50). Vision Transformers are competitive for fingerprints but lag on irises in this data regime, and cross-modal iris–fingerprint verification remains only slightly above chance. Feature visualizations indicate the models capture meaningful ridge and iris-texture patterns from low- to high-level representations. These findings motivate re-examining multi-trait fusion methods that assume independence and scaling cross-modal studies with larger datasets.

\clearpage
\newpage
{
    \small
    \bibliographystyle{ieeenat_fullname}
    \bibliography{strings}
}

\clearpage
\newpage

\appendix
\section{Appendix}

% --- Figure: Accuracy ---
\begin{figure}[ht]
  \centering
  \includegraphics[width=0.95\linewidth]{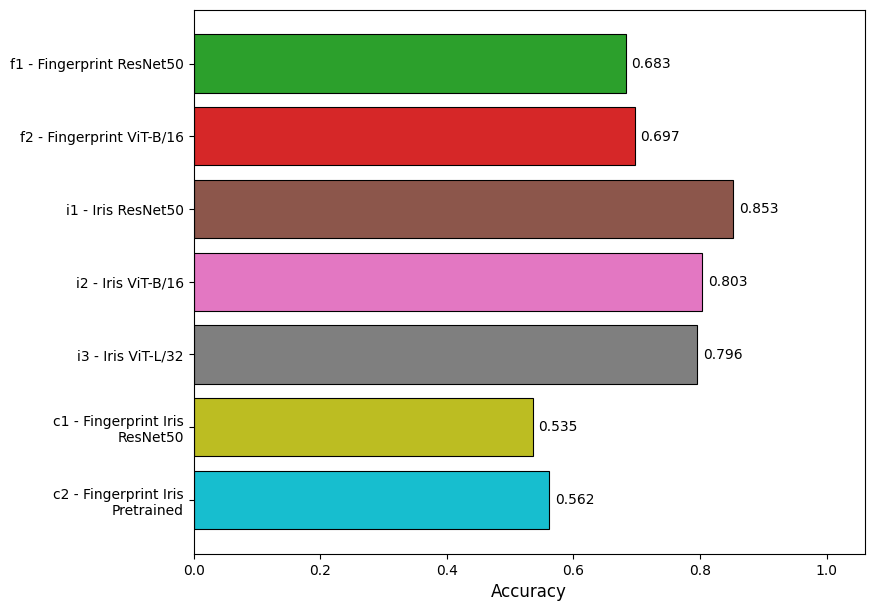}
  \caption{Accuracy on the held-out test set.}
  \label{fig:accuracy}
\end{figure}

% --- Figure: Precision ---
\begin{figure}[ht]
  \centering
  \includegraphics[width=0.95\linewidth]{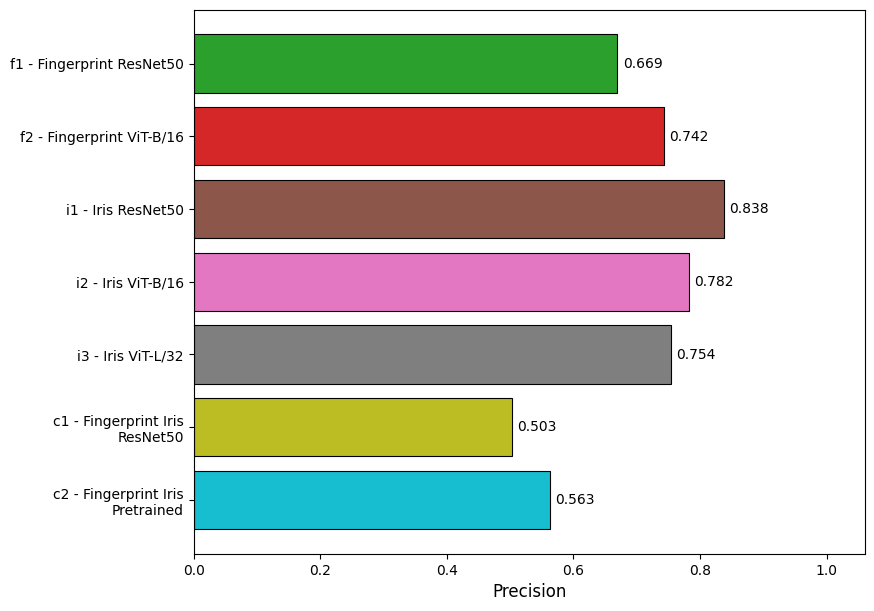}
  \caption{Precision on the held-out test set.}
  \label{fig:precision}
\end{figure}

% --- Figure: Recall ---
\begin{figure}[ht]
  \centering
  \includegraphics[width=0.95\linewidth]{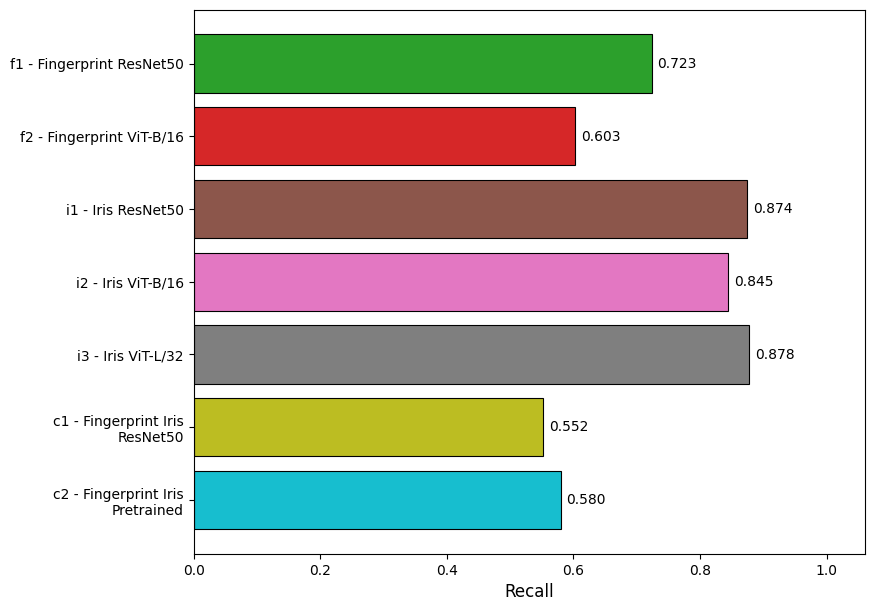}
  \caption{Recall on the held-out test set.}
  \label{fig:recall}
\end{figure}
\end{document}